\documentclass[10pt,journal,compsoc]{IEEEtran}
\usepackage[font=small, skip=0pt]{caption}
\usepackage{subcaption}
\usepackage{multicol}
\usepackage{graphicx}
\usepackage{amsmath}
\usepackage{tcolorbox}
\usepackage{booktabs}
\usepackage{url}
\usepackage[compact]{titlesec} 
\usepackage{xcolor}

\ifCLASSOPTIONcompsoc
  \usepackage[nocompress]{cite}
\else
  \usepackage{cite}
\fi
\ifCLASSINFOpdf
\else
\fi
\begin{document}

\definecolor{vividblue}{HTML}{0C7BDC}

\newcommand{\changed}[1]{\textcolor{black}{#1}}
    
\newcommand{\eg}{\hbox{e.g.,\ }}
\newcommand{\ie}{\hbox{i.e.,\ }}
\newcommand{\etal}{\hbox{\em et al.}}
\newcommand{\etals}{\hbox{\em et al.'s}}

\newcommand{\commentario}[1]{\textcolor{red}{\textbf{#1}}}
\newcommand{\ToDo}[1]{
    \textcolor{red}{[\textit{ToDo}: \textbf{#1}]}
    }
\newcommand{\XXX}{\textcolor{red}{XXX}\ }

\newcommand{\needsupdate}{\textcolor{red}{\textbf{This section has not been updated! Please don't read me! I'm old.}}\ }

\newcommand{\takeaway}[1]{\noindent
    \begin{tcolorbox}[]
    {#1}
    \end{tcolorbox}
}

\title{Lumen: A Machine Learning Framework to Expose Influence Cues in \changed{Texts}}

\author{Hanyu~Shi\textsuperscript{*}, 
    Mirela~Silva\textsuperscript{*}, 
	Daniel~Capecci,
	Luiz~Giovanini,
	Lauren~Czech, 
	Juliana~Fernandes,
	and Daniela~Oliveira
	
	\IEEEcompsocitemizethanks{
	\IEEEcompsocthanksitem H.~Shi, M.~Silva (corresponding author), D.~Capecci, L.~Giovanini, L.~Czech, and D.~Oliveira are with the Department of Electrical and Computer Engineering, University of Florida, Gainesville, FL, USA, 32611
	(E-mail: msilva1@ufl.edu).\protect
		
	\IEEEcompsocthanksitem J.~Fernandes is with the Department of Advertising, University of Florida, Gainesville, FL, USA, 326011.\protect
	
	}
	
	\thanks{* The first two authors have equal contribution.}
}

%

\IEEEtitleabstractindextext{%
\begin{abstract}
Phishing \changed{and disinformation are popular social engineering} attacks with attackers invariably applying influence cues in texts to make them more appealing to users. 
We introduce Lumen, a learning-based framework that exposes influence cues in text: (i) persuasion, (ii) framing, (iii) emotion, \changed{(iv) objectivity/subjectivity, (v) guilt/blame, and (vi) use of emphasis.}
\changed{Lumen was trained with a newly developed dataset of 3K texts comprised of disinformation, phishing, hyperpartisan news, and mainstream news.}
\changed{Evaluation of Lumen in comparison to other learning models showed that} Lumen and LSTM presented the best F1-micro score, but Lumen yielded better interpretability. Our results highlight the promise of ML to expose influence cues in text, \changed{towards the goal of application in automatic labeling tools to improve the} accuracy of human-based detection and reduce the likelihood of users falling for \changed{deceptive online content}.
\end{abstract}

	\begin{IEEEkeywords}
		Influence cues, machine learning, natural language processing, phishing, \changed{misinformation, deception detection}. 
\end{IEEEkeywords}}

\maketitle

\IEEEdisplaynontitleabstractindextext

%
\IEEEpeerreviewmaketitle

\IEEEraisesectionheading{\section{Introduction}\label{sec:intro}}

\IEEEPARstart{T}{he} \changed{Web has increasingly become an ecosystem for deception. Beyond social engineering attacks such as phishing which put Internet users and even national security at great peril~\cite{mueller2019report, Reuters_FactCheck_electionfraud-jk}, false information is greatly shaping the political, social, and economic landscapes of our society, exacerbated and brought to light in recent years by social media. Recent years have undoubtedly brought to light the dangers of selective exposure\footnotemark, and false content can increase individuals' beliefs in the falsehood~\cite{Ross2021-if}. These deceptive and divisive misuses of online media have evolved the previously seemingly tacit political lines to the forefront of our very own individual identities~\cite{Kalsnes2021-pg}, thus raising concern for the anti-democracy effects caused by this polarization of our society~\cite{Barnidge2019-tz}.
}

\footnotetext{\changed{A theory akin to \textit{confirmation bias} and often used in Communication research pertaining to the idea that individuals favor information that reinforces their prior beliefs~\cite{Stroud2014-vi}.}}

A key invariant of deceptive content is the application of influence cues in the text. Research on deception detection~\cite{Cialdini1993, Russell1980-og, Kircanski2018-rh, Rothman1997-jx, Kahneman1979-jz} reveals that deceivers apply influence cues in messages to increase their appeal to the recipients. We posit several types of influence cues that are relevant and prevalent in deceptive texts: (i) the principle of persuasion applied~\cite{Cialdini1993, Cialdini2001Science} (e.g., \textit{authority}, \textit{scarcity}), (ii) the framing of the message as either potentially causing a \textit{gain} or a \textit{loss}~\cite{Rothman1997-jx, Kahneman1979-jz}, (iii) the positive/negative emotional salience/valence of the content~\cite{Russell1980-og,Kircanski2018-rh},
\changed{
(iv) the subjectivity or objectivity of sentences in the text, (v) attribution of blame/guilt, and (vi) the use of emphasis.}

\changed{
Additionally, works such as Ross et al.~\cite{Ross2021-if} found that the ability to think deliberately and analytically (i.e., ``System 2''~\cite{Kahneman1979-jz}) is generally associated with the rejection of disinformation, regardless of the participants' political alignment---thus, the activation of this analytical thinking mode may act as an ``antidote'' to today's selective exposure.}
{\bf We therefore advocate that interventions should mitigate \changed{deceptive content} via the exposure of influence cues in texts.} 
\changed{
Similar to the government and state-affiliated media account labels on Twitter~\cite{Twitter_Help_Center_undated-lo}, bringing awareness to the influence cues present in misleading texts may, in turn, aid users by providing additional context in the message, thus helping users think analytically, and benefit future work aimed at the automatic detection of deceptive online content.
}

Towards this goal, we introduce Lumen\footnotemark, a two-layer learning framework that exposes influence cues in text using a novel combination of well-known existing methods: 
(i) topic modeling to extract structural features in text; (ii) sentiment analysis to extract emotional salience; 
\footnotetext{From Latin, meaning ``to illuminate.''} 
(iii) LIWC~\footnotemark to extract dictionary features related to influence cues; and (iv) a classification model to leverage the extracted features to predict the presence of influence cues. To evaluate Lumen's effectiveness, we leveraged our dataset of \changed{2,771 diverse pieces of online texts}, manually labeled by our research team according to the influence cues in the text using standard qualitative analysis methods. 
\changed{We must, however, emphasize that Lumen is \emph{not} a consumer-focused end-product, and instead is insomuch as a module for application in future user tools  that we shall make publicly available to be leveraged by researchers in future work (as described in Sec.~\ref{sec:limitations}).}

\footnotetext{\changed{LIWC~\cite{pennebaker2015development} is a transparent text analysis program that counts words in psychologically meaningful categories and is widely used to quantify psychometric characteristics of raw text data.}}

\changed{
Our newly developed dataset is comprised of nearly 3K texts, where 1K were mainstream news articles, and 2K deceptive or misleading content in the form of: Russia's Internet Research Agency's (IRA) propaganda targeting Americans in the 2016 U.S. Presidential Election, phishing emails, and fake and hyperpartisan news articles. 
Here, we briefly define these terms, which we argue fall within the same ``deceptive text umbrella.'' 
\emph{Disinformation} constitutes any any \emph{purposefully} deceptive content aimed at altering the opinion of or confusing an individual or group.
Within disinformation, we find instances of \emph{propaganda} (facts, rumors, half-truths, or lies disseminated manipulatively for the purpose of influencing public opinion~\cite{Smith2021-jq}) and \emph{fake news} (fabricated information that mimics real online news~\cite{Ross2021-if} and considerably overlaps with hyperpartisan news~\cite{Barnidge2019-tz}).
\textit{Mis}information's subtler, political form is \emph{hyperpartisan news}, which entails a misleading coverage of factual events through the lens of a strong partisan bias, typically challenging mainstream narratives~\cite{Ross2021-if, Barnidge2019-tz}.
\emph{Phishing} is a social engineering attack aimed at influencing users}
via deceptive arguments into an action (e.g., clicking on a malicious link) that will go against the user's best interests. 
\changed{
Though phishing differs from disinformation in its modus operandi, we argue that it overlaps with misleading media in their main purpose---to galvanize users into clicking a link or button by triggering the victim's emotions~\cite{Barnidge2019-tz}, and leveraging influence and deception.
}

\changed{
We conducted a quantitative analysis of the dataset, which showed that \textit{authority} and \textit{commitment} were the most common principles of persuasion in the dataset (71\% and 52\%, respectively), the latter of which was especially common in news articles. 
Phishing emails had the largest occurrence of \textit{scarcity} (65\%). 
Framing was a relatively rare occurrence (13\% \textit{gain} and 7\% \textit{loss}), though \textit{gain} framing was predominantly prevalent in phishing emails (41\%). The dataset invoked an overall positive sentiment (VADER compound score of $0.232$), with phishing emails containing the most positive average sentiment ($0.635$) and fake news with the most negative average sentiment ($-0.163$). 
\textit{Objectivity} and \textit{subjectivity} occurred in over half of the dataset, with \textit{objectivity} most prevalent in fake news articles (72\%) and \textit{subjectivity} most common in IRA ads (77\%). 
Attribution of blame/guilt was disproportionately frequent for fake and hyperpartisan news (between 38 and 45\%).
The use of emphasis was much more common in informal texts (e.g., IRA social media ads, 70\%), and less common in news articles (e.g., mainstream media, 17\%).
}

\changed{We evaluated Lumen in comparison with other traditional ML and deep learning algorithms.} Lumen presented the best performance in terms of its $F1$-micro score \changed{(69.23\%),
performing similarly to LSTM (69.48\%).}
In terms of $F1$-macro, LSTM \changed{(64.20\%)} performed better than Lumen \changed{(58.30\%)}; however, Lumen presented better interpretability for intuitively understanding of the model, as it provides both the relative importance of each feature and the topic structure of the training dataset without additional 
computational costs, which cannot be obtained with LSTM as it operates as a black-box. 
Our results highlight the promise of exposing influence cues in text via learning methods. 


    

\footnotetext{\changed{Lumen and dataset will be available upon publication.}}

This paper is organized as follows. Section~\ref{sec:relatedwork} positions this paper's contributions in comparison to related work in the field. Section~\ref{sec:method_dataset_coding} details the methodology used to generate our coded dataset. 
Section~\ref{sec:design-impl} describes Lumen's design and implementation, as well as Lumen's experimental evaluation.
Section~\ref{sec:eval} contains a quantitative analysis of our dataset, and Lumen's evaluation and performance. Section~\ref{sec:discussion} summarizes our findings and discusses the limitations of our work, \changed{as well as recommendations for future work.} 
Section~\ref{sec:conclusion} concludes the paper.

\section{Related Work}
\label{sec:relatedwork}

This section briefly summarizes the extensive body of work on machine learning methods to automatically 
\changed{detect disinformation and hyperpartisan news}, and initial efforts to detect the presence of influence cues in text. 

\subsection{Automatic Detection of \changed{Deceptive Text}}

\subsubsection{\changed{Phishing and Spam}}
Most anti-phishing research has focused on automatic detection of malicious messages and URLs before they reach a user's inbox via a combination of blocklists~\cite{Dong2015-sm,Oest2019-jk} and ML~\cite{Peng2018-jg, Bursztein2019-hq}. 
Despite yielding high filtering rates in practice, these approaches cannot prevent zero-day phishing\footnotemark from reaching users because determining maliciousness of text is an open problem and phishing constantly changes, rendering learning models and blocklists outdated in a short period of time~\cite{Bursztein2019-hq}.
Unless the same message has been previously reported to an email provider as malicious by a user or the provider has the embedded URL in its blocklist, determining maliciousness is extremely challenging. 
Furthermore, the traditional approach to automatically detect phishing takes a binary standpoint (phishing or legitimate, e.g., \cite{Basnet2008, Shyni2016multi, chandrasekaran2006phishing}), potentially overlooking distinctive nuances and the sheer diversity of malicious messages. 

\footnotetext{\changed{A new, not-yet reported phishing email.}}

Given the limitations of automated detection in handling zero-day phishing, human detection has been proposed as a complementary strategy. The goal is to either warn users about issues with security indicators in web sites, which could be landing pages of malicious URLs~\cite{Felt2015-wy, Sunshine2009-kr} or train users into recognizing malicious content online~\cite{Sheng2007-jn}. These approaches are not without their own limitations. 
For example, research on the effectiveness of SSL warnings shows that users either habituate or tend to ignore warnings due to false positives or a lack of understanding about the warning message~\cite{Vance2017-oe, Akhawe2013-ha}. 
\subsubsection{\changed{Fake \& Hyperpartisan Media}}

\changed{The previously known ``antidote'' to reduce polarization and increase readers' tolerance to selective exposure was via the use of counter-dispositional information~\cite{Barnidge2019-tz}. However, countering misleading texts with mainstream or high-quality content in the age of rapid-fire social media comes with logistical and nuanced difficulties. Pennycook and Rand~\cite{Pennycook2021-sh} provide a thorough review of the three main approaches employed in fighting misinformation: automatic detection, debunking by field experts (which is not scalable), and exposing the publisher of the news source.}

\changed{
Similar to zero-day phishing, disinformation is constantly morphing, such that ``zero-day'' disinformation may thwart already-established algorithms, such was the case with the COVID-19 pandemic~\cite{Pennycook2021-sh}.}
\changed{Additionally, the final determination of a \emph{fake, true}, or \emph{hyperpartisan} label is fraught with subjectivity. Even fact-checkers are not immune---their agreement rates plummet for ambiguous statements~\cite{Lim2018-ic}, calling into question their efficacy in hyperpartisan news.} 




\changed{We posit that one facet of the solution lies within the combination of human and automated detection. Pennycook and Rand~\cite{Pennycook2021-sh} conclude that lack of careful reasoning and domain knowledge is linked to poor truth discernment, suggesting (alongside \cite{Ross2021-if, Bago2020-wc}) that future work should aim to trigger users to to think slowly and analytically~\cite{Kahneman1979-jz} while assessing the accuracy of the information presented.}
\changed{Lumen aims to fulfill the first step of this goal, as our framework exposes influence cues in texts,  which we hypothesize is disproportionately leveraged in deceptive content.} 


\subsection{Detecting \changed{Influence in Deceptive Texts}}

\subsubsection{\changed{Phishing}}
We focus on prior work that has investigated the extent to which Cialdini's principles of persuasion (PoP)~\cite{Cialdini2001Science, Cialdini1993} \changed{(described in Sec.~\ref{sec:method_dataset_coding})} are used in phishing emails~\cite{Stajano2011-el, Oliveira2019-wq, Ferreira2019-tr} and how users are susceptible to them~\cite{Oliveira2017dissect, Lawson2017-rn}. 

\changed{Lawson et al.~\cite{Lawson2017-rn} leveraged a personality inventory and an email identification task to investigate the relationship between personality and Cialdini's PoP. 
The authors found that \emph{extroversion} was significantly correlated with increased susceptibility to \emph{commitment, liking,} and the pair \emph{(authority, commitment)}, the latter of which was found in 41\% of our dataset.}
Following Cialdini's PoP, after manually labeling $\sim$200 phishing emails, Akbar~\cite{akbar2014analysing} found that \emph{authority} was the most frequent principle in the phishing emails, followed by \emph{scarcity}, corroborating our findings for high prevalence of \emph{authority}.
However, in a large-scale phishing email study with more than 2,000 participants, Wright et al.~\cite{Wright2014} found that \emph{liking} receives the highest phishing response rate, while \emph{authority} received the lowest. Oliveira et al.~\cite{Oliveira2017dissect, Lin2019Susceptibility} unraveled the complicated relationship between PoP and Internet user age and susceptibility, finding that young users are most vulnerable to \emph{scarcity}, while older ones are most likely to fall for \emph{reciprocation}, with \emph{authority} highly effective for both age groups. 
\changed{These results are promising in highlighting the potential usability of exposing influence cues to users.}
\subsubsection{\changed{Fake \& Hyperpartisan News}}
\changed{Contrary to phishing, few studies have focused on detecting influence cues or analyzing how users are susceptible to them in the context of fake or highly partisan content. 
Xu et al.~\cite{Xu2020-fm} stands out as the authors used a mixed-methods analysis, leveraging both manual analysis of the textual content of 1.2K immigration-related news articles from 17 different news outlets, and computational linguistics (including, as we did, LIWC). The authors found that moral frames that emphasize that support \emph{authority/respect} were shared/liked more, while the opposite occurred for \emph{reciprocity/fairness}. Whereas we solely used trained coders, they measured the aforementioned frames by applying the moral foundations dictionary~\cite{Graham2009-ns}.}


To the best of our knowledge, no prior work has investigated or attempted to automatically detect influence cues in texts \changed{in such a large dataset, containing multiple types of deceptive texts.} In this work, we go beyond Cialdini's principles to also detect gain and loss framing, emotional salience, \changed{subjectivity and objectivity, and the use of emphasis and blame.} Further, no prior work has made available to the research community a dataset of \changed{deceptive texts labeled according to the influence cues} applied in the text.

\section{\changed{Dataset Curation \& Coding Methodology}}
\label{sec:method_dataset_coding}
\changed{This section describes the methodology to generate the labeled dataset of online texts used to train Lumen, including the definition of each of the influence cues labels.}

\subsection{\changed{Curating the Dataset}}
\changed{We composed a diverse dataset by gathering different types of texts from multiple sources, split into three groups: ({\bf Deceptive Texts}) $1,082$ pieces of text containing disinformation and/or deception tactics, ({\bf Hyperpartisan News)} $1,003$ hyperpartisan media news from politically right- and left-leaning publications, and ({\bf Mainstream News}) $974$ center mainstream media news. Our dataset therefore contained $3,059$ pieces of text in total.}

\changed{For the \textbf{Deceptive Texts Group}, we mixed $492$ Facebook ads created by the Russian Internet Research Agency (IRA), $130$ known fake news articles, and $460$ phishing emails:}

\changed{{\bf Facebook IRA Ads.} We leveraged a dataset of 3,517 Facebook ads created by the Russian IRA and made publicly available to the U.S. House of Representatives Permanent Select Committee on Intelligence~\cite{noauthor_undated-wp} by Facebook after internal audits. These ads were a small representative sample of over 80K organic content identified by the Committee and are estimated to have been exposed to over 126M Americans between June 2015 and August 2017. After discarding ads that did not have text entry, the dataset was decreased to 3,286 ads, which were mostly (52.8\%) posted in 2016 (U.S. election year).
We randomly selected $492$ for inclusion.}

\changed{{\bf Fake News.} We leveraged a publicly available\footnotemark dataset of nearly 17K news labeled as \emph{fake} or \emph{real} collected by Sadeghi et al.~\cite{fbzd-sw81-20} from \url{PoliticFact.com}, a reputable source of fact-finding. 
We randomly selected $130$ \emph{fake} news ranging from $110-200$ words dated between 2007 to 2020.}

\footnotetext{\changed{\url{https://ieee-dataport.org/open-access/fnid-fake-news-inference-dataset#files}}}

\changed{{\bf Phishing Emails.}} To gather our  dataset, we collected approximately 15K known phishing emails  from multiple public sources~\cite{Smiles2019Phishing, Pittsburgh2019Alerts, Minnesota2019Phishing, UCLA2019Phish, Penn2019Phish, Lehigh2019Recent, UA2019Phishing, Michigan2019Phish}. The emails were then cleaned and formatted to remove errors, noise (e.g., images, HTML elements), and any extraneous formatting so that only the raw email text remained. \changed{We randomly selected $460$ of these emails ranging from $50-150$ words to be included as part of the Deceptive Texts.}

\changed{For the \textbf{Hyperpartisan News} and \textbf{Mainstream News Groups}, we used a public dataset\footnotemark comprised of 2.7M news articles and essays from 27 American publications dated from 2013 to early 2020.
\footnotetext{\changed{\url{https://components.one/datasets/all-the-news-2-news-articles-dataset/}}}
We first selected articles ranging from $50-200$ words and then classified them as \emph{left}, \emph{right}, or \emph{center} news according to the AllSides Bias Rating\footnotemark. For inclusion in  the Hyperpartisan News Group, we randomly selected $506$ \emph{right} news and $497$ \emph{left} news; the former were dated from 2016 to 2017 and came from two publications sources (Breitbart and National Review) while the latter were dated from 2016 to 2019 and came from six publications (Buzzfeed News, Mashable, New Yorker, People, VICE, and Vox). To compose Mainstream News Group, we randomly selected $974$ \emph{center} news from all seven publications (Business Insider, CNBC, NPR, Reuters, TechCrunch, The Hill, and Wired) dated from 2014 to 2019.}

\footnotetext{\changed{\url{https://www.allsides.com/media-bias/media-bias-ratings}}}

\subsubsection{\changed{Coding Process}}
\changed{We then developed coding categories and a codebook based on Cialdiani’s principles of influence~\cite{cialdini1993psychology}, subjectivity/objectivity, and gain/loss framing~\cite{kahneman1980prospect}. These categories have been used in prior works (e.g., \cite{Oliveira2019-wq, Oliveira2017dissect}, and were adapted for the purposes of this study, as well as with the additional emphasis and blame/guilt attribution categories. Next, we held an initial training session with nine undergraduate students. The training involved a thorough description of the coding categories, their definitions and operationalizations, as well as a workshop-style training where coders labeled a small sample of the texts to get acquainted with the coding platform, the codebook, and the texts. Coders were instructed to read the text at least twice before starting the coding to ensure they understood it. After that, coders were asked to share their experiences labeling the texts and to discuss any issues or  questions about the process. After this training session, two intercoder reliability pretests were conducted; in the first pretest, coders independently co-coded a sample of $20$ texts, and in the second pretest, coders independently co-coded a sample of $40$ texts. After each one of these pretests, a discussion and new training session followed to clarify any issues with the categories and codebook.}

\changed{Following these additional discussion and training sessions, coders were then instructed to co-code $260$ texts which served as our intercoder reliability sample. To calculate intercoder reliability, we used three indexes. \emph{Cohen’s kappa} and \emph{Percent of Agreement} ranged from $0.40$ to $0.90$, and 66\% to 99\%, respectively, which was considered moderately satisfactory. Due to the nature of the coding and type of texts, we also opted to use \emph{Perrault and Leigh’s index} because (a) it has been used in similar studies that also use nominal data~\cite{fuller2014lights, hove2013newspaper, morey2016measures, rice2018frequent}; (b) it is the most appropriate reliability measure for $0/1$ coding (i.e., when coders mark for absence or presence of given categories), as traditional approaches do not take into consideration two zeros as agreement and thus penalize reliability even if coders disagree only a few times~\cite{perreault1989reliability}; and (c) indexes such as Cohen’s kappa and Scott’s pi have been criticized for being overly conservative and difficult to compare to other indexes of reliability~\cite{lombard2002content}. Perrault and Leigh’s index ($I_r$) returned a range of $0.67$ to $0.99$, which was considered satisfactory. Finally, the remaining texts were divided equally between all coders, who coded all the texts independently using an electronic coding sheet in Qualtrics. 
Coders were instructed to distribute their workload equally over the coding period to counteract possible fatigue effects. This coding process lasted three months.}

\subsubsection{\changed{Influence Cues Definitions}}
\changed{The coding categories were divided into five main concepts: \emph{principles of influence}, \emph{gain/loss framing},  \emph{objectivity/subjectivity}, \emph{attribution of guilt}, and \emph{emphasis}. Coders marked for the absence ($0$) or presence ($1$) of each of the categories. Definitions and examples for each influence are detailed in Appendix~A, leveraged from the coding manual we curated to train our group of coders.}

\changed{{\bf Principles of Persuasion (PoP).}} \emph{Persuasion} refers to a set of principles that influence how people concede or comply with requests. 
\changed{The principles of influence were based on Cialdini’s marketing research work~\cite{Cialdini1993,Cialdini2001Science}, and consist of the following six principles: (i) \emph{authority\footnotemark} or expertise, (ii) \emph{reciprocation}, (iii) \emph{commitment} and consistency, (iv) \emph{liking}, (v) \emph{scarcity}, and (vi) \emph{social proof}. We added subcategories to the principles of commitment (i.e., indignation and call to action) and social proof (i.e., admonition) because an initial perusal of texts revealed consistent usage across texts.}

\footnotetext{e.g., people tend to comply with requests or accept arguments made by figures of authority.}

\changed{{\bf Framing.}} Framing refers to the presentation of a message (e.g., health message, financial options, and advertisement)  as implying a possible gain (i.e., possible benefits of performing the action) vs. implying a possible loss (i.e., costs of not performing a behavior)~\cite{Rothman1997-jx,Kuhberger1998-cs,Kahneman1979-jz}. 
Framing can affect decision-making and behavior; work by Kahneman and Tversky~\cite{ Kahneman1979-jz} on loss aversion supports the human tendency to prefer avoiding losses over acquiring equivalent gains.

\changed{{\bf Slant.} Slant refers to whether a text is written subjectively or objectively; \emph{subjective}sentences generally refer to a personal opinion/judgment or emotion, whereas \emph{objective} sentences fired to factual information that is based on evidence, or when evidence is presented. It is important to note that we did not ask our coders to fact check, instead asking them to rely on sentence structure, grammar, and semantics to determine the label of \emph{objective} or \emph{subjective}.}

\changed{{\bf Attribution of Blame/Guilt.} Blame or guilt refers to when a text references “another” person/object/idea for wrong or bad things that have happened.}

\changed{{\bf Emphasis.} Emphasis refers to the use of all caps text, exclamation points (either one or multiple), several question marks, bold text, italics text, or anything that is used to call attention in text.}

\section{Lumen Design and Implementation}
\label{sec:design-impl}

This section describes the design, implementation and evaluation of Lumen, our proposed two-level learning-based framework to expose influence cues in texts.

\subsection{\changed{Lumen Overview}}
Exposing presence of persuasion and framing is tackled as a multi-labeling document classification problem, where zero, one, or more labels can be assigned to each document. Due to recent developments in natural language processing, emotional salience is an input feature that Lumen exposes leveraging sentiment analysis.  Note that  Lumen's goal is not to distinguish \changed{deceptive vs. benign texts}, but to expose different influence cues applied in different types of texts.

\begin{figure*}[h]
	\centering
		\includegraphics[width=0.8\textwidth,trim={0 7.2cm 3cm 1.3cm}, clip]{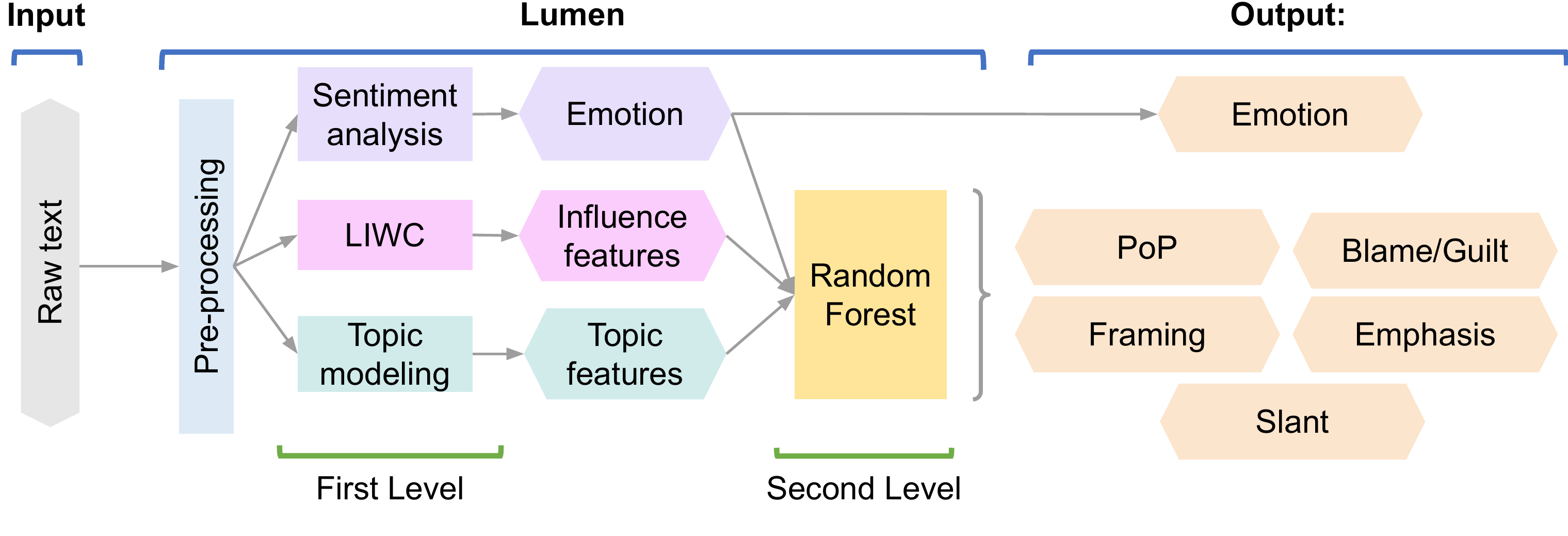}
	\caption{Lumen's two-level architecture. Pre-processed text undergoes sentiment analysis for extraction of emotional salience, LIWC analysis for extraction of features related to influence keywords, and topic modeling for structural features. These features are inputs to ML analysis for prediction of influence cues applied to the message.}
	\label{fig:lumen}
\end{figure*}

Figure~\ref{fig:lumen} illustrates Lumen’s two-level hierarchical learning- based architecture. On the first level, the following features are extracted from the raw text: (i) topical structure inferred by topic modeling, (ii) LIWC features related to influence keywords~\cite{pennebaker2015development}, and (iii) emotional salience features learned via sentiment analysis~\cite{Hutto2014}. On the second level, a classification model is used to identify the influence cues existing in the text.

\subsection{Topic Structure Features} 
Probabilistic topic modeling algorithms are often used to infer the topic structure of unstructured text data~\cite{Steyvers2007,BleiBook2009}, which in our case are \changed{deceptive texts, hyperpartisan news, and mainstream news.} Generally, these algorithms assume that a collection of documents (i.e., a corpus) are created following the generative process. 

Suppose that there are $D$ documents in the corpus $\mathcal{C}$ and each document $d=1,...,D$ has length $m_d$. Also suppose that there are in total $K$ different topics in the corpus and the vocabulary $\mathcal{V}$ includes $V$ unique words. The relations between documents and topics are determined by conditional probabilities $P(t|d)$, which specify the probability of topic $t=1,...,K$ given document $d$. The linkage between topics and unique words are established by conditional probabilities $P(w|t)$, which indicate the probability of word $w=1,...,V$ given topic $t$. According to the generative process, for each token $w(i_d)$, which denotes the $i_d$-th word in document $d$, we will first obtain the topic of this token, $z(i_d)=t$, according to $P(t|d)$. With the obtained $z(i_d)$, we then draw the a word $w(i_d)=w$ according to $P(w|t=z(i_d))$ .

In this work, we leveraged Latent Dririchlet Allocation (LDA), one of the most widely used topic modeling algorithms, to infer topic structure in texts~\cite{Blei2003lda}. In LDA, both $P(w|t)$ and $P(t|d)$ are assumed to have Dirichlet prior distributions. Given our dataset, which is the evidence to the probabilistic model, the goal of LDA is to infer the most likely conditional distribution $\hat P(w|t)$ and $\hat P(t|d)$, which is usually done by either variational Bayesian approach~\cite{Blei2003lda} or Gibbs Sampling~\cite{Griffiths2004}. In Lumen, the conditional probabilities $\hat P(t|d)$ represent the topic structure of the dataset.

\subsection{LIWC Influence Features} 

We use language to convey our thoughts, intentions, and emotions, with words serving as the basic building blocks of languages. Thus, the way different words are used in unstructured text data  provide meaningful information to streamline our understanding of the use of influence cues in text data. Lumen thus leverages LIWC, a natural language processing framework that connects commonly-used words with categories~\cite{tausczik2010psychological, pennebaker2015development} to retrieve influence features of texts to aid ML classification. LIWC includes more than 70 different categories in total, such as Perceptual Processes, Grammar, and Affect, and more than 6K common words.

However, not all the categories are related to influence. After careful inspection, we manually selected seven  categories as features related to influence for Lumen. For persuasion, we selected the category \emph{time} (related to \emph{scarcity}); for emotion, we selected the categories \emph{anxiety, anger,} and \emph{sad}; and for framing, we selected the categories \emph{reward} and \emph{money} (gain), and \emph{risk} (loss). 


We denote the collection of the chosen LIWC  categories as set $\mathcal{S}$. Given a text document $d$ with document length $m_d$ from the corpus $\mathcal{C}$, to build the LIWC  feature $X^{LIWC}_{i,d}, \forall  i \in \mathcal{S}$,  we first count the number of words in the text $d$ belonging to the LIWC category $i$, denoted as $n_{i,d}$, then normalize the raw word count with the document length:
\begin{equation}
X^{LIWC}_{i,d}=\frac{n_{i,d}}{m_d}, \forall  i \in \mathcal{S}, d \in \mathcal{C}.
\end{equation}

\subsection{Emotional Salience by Sentiment Analysis}
Emotional salience refers to both valence (positive to negative) and arousal (arousing to calming) of an experience or stimulus~\cite{Russell1980-og,Peace2012-dm,Kircanski2018-rh}, and research has shown that deception detection is reduced for emotional compared to neutral stimuli~\cite{Peace2012-dm}. Similarly, persuasion messages that generate high (compared to low) arousal lead to poorer consumer decision-making~\cite{Kircanski2018-rh}. Emotional salience may impair full processing of deceptive content and high arousal content may trigger System 1, the fast, shortcut-based brain processing mode~\cite{Ariely2009-op}.

In this work, we used a pre-trained rule-based model, VADER, to extract the emotional salience and valence from a document~\cite{Hutto2014}. 
Both levels of emotion range from 0 to 1, where a small value means low emotional levels and a large number means high emotional levels. Therefore, emotional salience is both an input feature to the learning model and one of Lumen’s outputs (see Fig.~\ref{fig:lumen}).

\subsection{Machine Learning to Predict Persuasion \& Framing}
Lumen's second level corresponds to the application of a general-purpose ML algorithm for document classification. Although Lumen is general enough to allow application of any general-purpose algorithm, in this paper, we applied Random Forest (RF) because it can provide the level of importance for each input predicative feature without additional computational cost, which aids in model understanding. Another advantage of RF is its robustness to the magnitudes of input predicative features, i.e., RF does not need feature normalization. \changed{We use the grid search approach to fine-tune the parameters in the RF model and follow the cross-validation to overcome any over-fitting issues of the model.}



\subsubsection{Dataset Pre-Processing}
As described previously, Lumen generates three types of features at its first hierarchical level (emotional salience, LIWC categories, and topic structure), which serve as input  for the learning-based prediction algorithm (Random Forest, for this analysis) at Lumen's second hierarchical level (Fig.~\ref{fig:lumen}); these features rely on the unstructured texts in the dataset. However, different features need distinct preprocessing procedures. In our work, we used the Natural Language Toolkit (NLTK)~\cite{nlt} to pre-process the dataset. For all three types of features, we first removed all the punctuation, special characters, digital numbers, and words with only one or two characters. Next, we tokenized each document into a list of lowercase words.

For topic modeling features, we removed stopwords (which provide little semantic information) and applied stemming (replacing a word's inflected form with its stem form) to further clean-up word tokens. For LIWC features, we matched each word in each text with the pre-determined word list in each LIWC category; we also performed stemming for LIWC features. We did not need to perform pre-processing for emotional salience because we applied NLTK~\cite{Hutto2014}, which has its own tokenization and pre-processing procedures.

Additionally, we filtered out documents with less than ten words since topic modeling results for extremely short documents are not reliable~\cite{shi2019evaluation}. \changed{We were then left with 2,771 cleaned documents, with 183,442 tokens across the corpus, and 14,938 unique words in the vocabulary}.

\subsubsection{Training and Testing} 
Next, we split the \changed{the 2,771 documents} into a training and a testing set. 
In learning models, hyper-parameters are of crucial importance because they control the structure or learning processing of the algorithms. Lumen applies two learning algorithms: an unsupervised topic modeling algorithm, LDA, on the first hierarchical level and RF on the second level. Each algorithm introduces its own types of hyper-parameters; for LDA, examples include the number of topics and the concentration parameters for Dirichlet distributions, whereas for RF are the number of trees and the maximum depth of a tree. We also used the grid search approach to find a better combination of hyper-parameters. Note that due to time and computational power constraints, it is impossible to search all hyper-parameters and all their potential values. In this work, we only performed the grid search for number of topics (LDA) and the number of trees (RF). The results show that the optimal number of topics is 10 and the optimal number of trees in RF is 200. Note also that the optimal result is limited by the grid search space, which only contains a finite size of parameter combinations.

If we only trained and tested Lumen on one single pair of training and testing sets, there would be high risk of overfitting. To lower this risk, we used 5-fold cross-validation, wherein the final performance of the learning algorithm is the average performance over the five training and testing pairs.

\subsubsection{Evaluation Metrics}
To evaluate our results (Sec.~\ref{sec:eval}), we compared Lumen's performance in predicting the influence cues applied to a given document with three other document classification algorithms: (i) Labeled-LDA, (ii) LSTM, and (iii) naïve algorithm.

\changed{{\bf Labeled-LDA}} is a semi-supervised variation of the original LDA algorithm~\cite{Ramage2009, VanderHeijden2019}. When training the Labeled-LDA, both the raw document and the human coded labels for influence cues were input into the model. Compared to Lumen, Labeled-LDA only uses the word frequency information from the raw text data and has a very rigid assumption of the relation between the word frequency information and the coded labels, which limits its flexibility and prediction ability.

\changed{{\bf Long Short-Term Memory}} (LSTM) takes the input data recurrently, regulates the flow of information, and determines what to pass on to the next processing step and what to forget. Since neural networks mainly deal with vector operations, we used \changed{50-dimensional} word embedding matrix to transfer each word into vector space~\cite{NAILI2017340}. The main shortcoming of neural network is that it works as a blackbox, making it difficult to understand the underlying mechanism.

\changed{The {\bf na\"{\i}ve algorithm}} served as a base line for our evaluation. We randomly generated each label for each document according to a Bernoulli distribution with equal probabilities for two outcomes.

As shown in Table~\ref{tab:learning_results}, we used $F1$-score (following the work by Ramage et al.~\cite{Ramage2009} and van der Heijden et al.~\cite{VanderHeijden2019}), \changed{and accuracy rate} to quantify the performance of the algorithms. 
We note that the comparison of $F$-scores is only meaningful under the same experiment setup. It would be uninformative to compare F-scores from distinctive experiments in different pieces of work in the literature due to varying experiment conditions. $F1$-score can be easily calculated for single-labeling classification problems, where each document will only be assigned to one label. However, in our work, we are dealing with a multi-labeling classification problem, which means that no limit is imposed on how many labels each document can include. Thus, we employed two variations of the $F1$-score to quantify the overall performance of the learning algorithm: macro and micro $F1$-scores.


\section{Results}
\label{sec:eval}

This section details Lumen’s evaluation. First, we provide a quantitative analysis of our newly developed dataset used to train Lumen, and the results of Lumen classification in comparison to other ML algorithms.

\subsection{Quantitative Analysis of the Dataset}
\changed{We first begin by quantifying the curated dataset of 2,771 deceptive, hyperpartisan, or mainstream texts, hand-labeled by a group of coders. When considering all influence cues, most texts used between three and six cues per texts; only 3\% of all texts leveraged a single influence cue, and 2\% used zero cues ($n=58$).}

\changed{When considering the most common pairs and triplets between all influence cues, slant (i.e., \emph{subjectivity} or \emph{objectivity}) and principles of persuasion (PoP) dominated the top 10 most common pairings and triplets. As such, the most common pairs were \emph{(authority, objectivity)} and \emph{(authority, subjectivity)}, occurring for 48\% and 45\% of all texts, respectively. The most common (PoP, PoP) pairing was between \emph{authority} and \emph{commitment}, co-occurring in 41\% of all texts. \emph{Emphasis} appeared once in the top 10 pairs and twice in the top triplets: \emph{(emphasis, subjectivity)} occurring for 29\% of texts, and \emph{(emphasis, authority, subjectivity)} and \emph{(emphasis, commitment, subjectivity)} for 20\% and 19\% of texts, respectively. \emph{Blame/guilt} appeared only once in  the top triplets as \emph{(authority, blame/guilt, objectivity)}, representing 19\% of all texts. \emph{Gain} framing appeared only as the 33rd most common pair \emph{(gain, subjectivity)} and 18th most common triplet \emph{(call to action, scarcity, gain)}, further emphasizing its scarcity in our dataset.}

\subsubsection{\changed{Principles of Persuasion}}
\begin{figure}[h]
	\centering
	\includegraphics[width=0.5\textwidth]{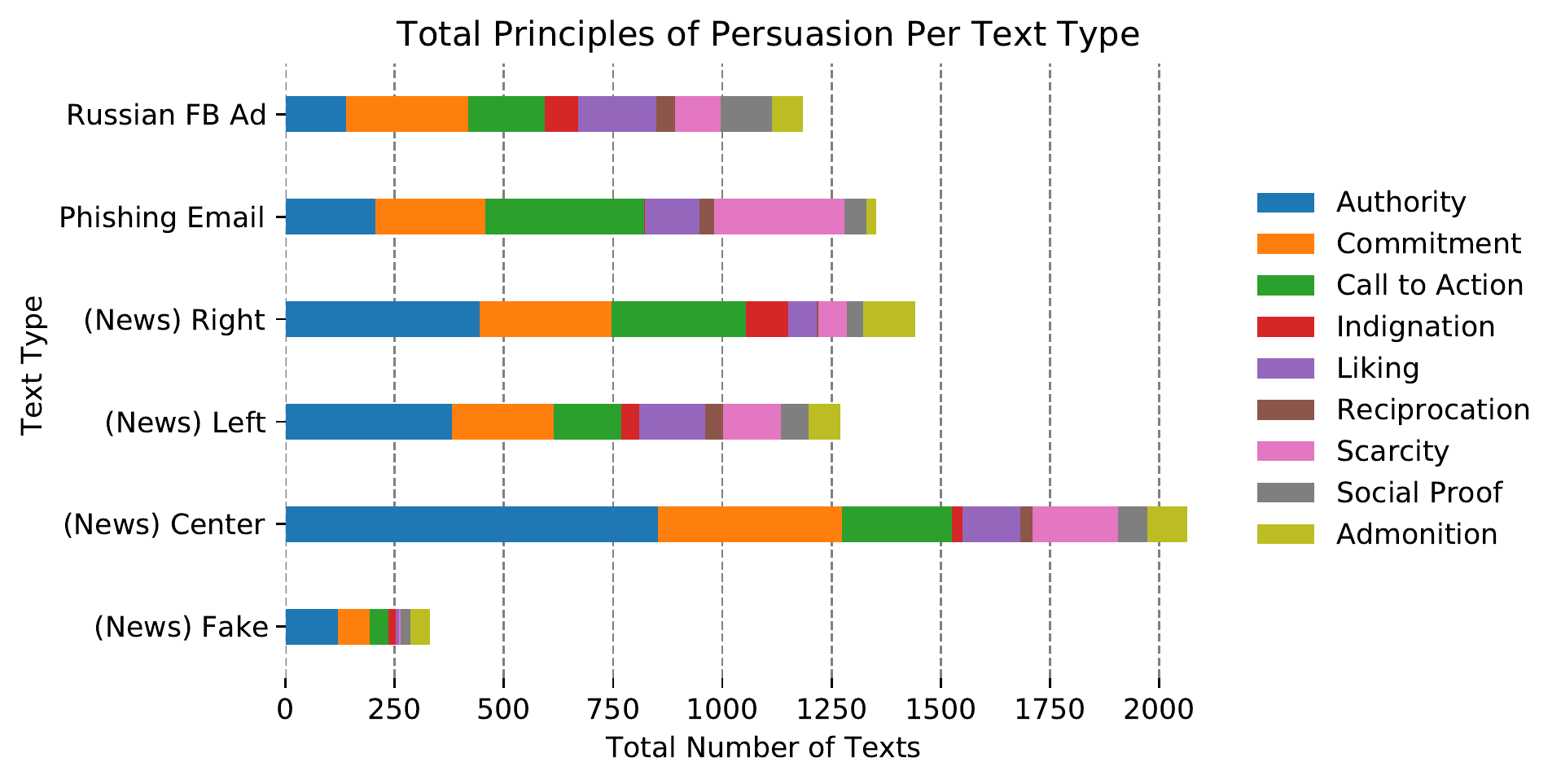},
	\caption{\changed{The total for each text type, broken down based on principles of persuasion.}}
    \label{fig:pop_breakdown_per_type}
\end{figure}

\changed{We found that most texts in the dataset contained one to four principles of persuasion, with only 4\% containing zero and 3\% containing six or more PoP labels; 29\% of texts apply two PoP and 23\% leverage three PoP. Further, Fig.~\ref{fig:pop_breakdown_per_type} shows that \emph{authority} and \emph{commitment} were the most prevalent principles appearing, respectively, in 71\% and 52\% of the texts; meanwhile, \emph{reciprocation} and \emph{indignation} were the least common PoP (5\% and 9\%, respectively).}


\changed{Almost all types of texts contained every PoP to varying degrees; the only exception is \emph{reciprocation} (the least-used PoP overall) which was not at all present in fake news texts (in the Deceptive Texts Group) and barely present ($n=3$, 0.6\%) in right-leaning hyperpartisan news. \emph{Authority} was the most-used PoP for all types of texts, except phishing emails (most: \emph{call to action}) and IRA ads (most: \emph{commitment}), both of which are in the Deceptive Texts Group.}


\changed{{\bf Deceptive Texts.} Fake news was notably reliant on \emph{authority} (92\% of all fake news leveraged the \emph{authority} label) compared to phishing emails (45\%) and the IRA ads (32\%); however, fake news used \emph{liking, reciprocation}, and \emph{scarcity} (5\%, 0\%, 3\%, respectively) much less often than phishing emails (27\%, 8\%, 65\%) or IRA ads (41\%, 10\%, 24\%). Interestingly, \emph{admonition} was most used by fake news (35\%), though overall, \emph{admonition} was only present in 14\% of all texts. Phishing emails were noticeably more reliant on \emph{call to action} (80\%) and \emph{scarcity} (65\%) compared to fake news (33\%, 3\%) and IRA ads (40\%, 24\%), yet barely used \emph{indignation} (0.4\%) compared to the same (13\% for fake news and 17\% for IRA ads). The IRA ads relied on \emph{indignation, liking, reciprocation}, and \emph{social proof} much more than the others; note again that \emph{reciprocation }was the least occurring PoP (5\% overall), but was most commonly occurring in IRA ads (10\%). }

\changed{{\bf Hyperpartisan News.} Right-leaning texts had nearly twice as much \emph{call to action} and \emph{indignation} than left-leaning texts (61\% and 19\% vs. 31\% and 8\%, respectively). Meanwhile, left-leaning hyperpartisan texts had noticeably more \emph{liking} (30\% vs. 13\%), \emph{reciprocation} (8\% vs. 0.6\%), and \emph{scarcity} (27\% vs. 13\%) than right-leaning texts. 
}

\changed{{\bf Mainstream News.} \emph{Authority} (88\%) and \emph{commitment} (43\%) were the most frequently appearing PoP in center news, though this represents the third highest occurrence of \emph{authority} and lowest use of \emph{commitment} across all six text type groups.  Mainstream news also used very little \emph{indignation} (3\%) compared the the other text types except phishing emails (0.4\%), and also demonstrated the lowest use of \emph{social proof} (7\%).}


\takeaway{\changed{\emph{Authority} and \emph{commitment} were the most common PoP in the dataset, with the former most common in fake news articles. Phishing emails had the largest occurrence of \emph{scarcity}).}}

\subsubsection{\changed{Framing}}
\changed{There were few \emph{gain} or \emph{loss} labels for the overall dataset (only 13\% and 7\%, respectively). Very few texts (18\%) were framed exclusively as either \emph{gain} or \emph{loss}, 81\% did not include any framing at all, and only 1\% of the texts used both \emph{gain} and \emph{loss} framing in the same message. We also found that \emph{gain} was much more prevalent than \emph{loss} across all types of texts, except for fake news, which showed an equal amount (1.5\% for both \emph{gain} and \emph{loss}). Notably, phishing emails had significantly more \emph{gain} and \emph{loss} than any other text type (41\% and 29\%, respectively); mainstream center news and IRA ads showed some use of \emph{gain} framing (10\% and 13\%, respectively) compared to the remaining text types.}

\changed{Next, we investigated how persuasion and framing were used in texts by analyzing the pairs and triplets between the two influence cues. \emph{Gain} framing most frequently occurred with \emph{call to action} and \emph{commitment}, though these represent only 9\% of pairings. \emph{Gain, call to action} and \emph{scarcity} was the the most common triplet between PoP and framing, occurring for 7\% of all texts—this is notable as phishing emails had \emph{call to action} and \emph{scarcity} as its top PoP, and \emph{gain} framing was also most prevalent in phishing. Also of note is that \emph{loss} appeared in even fewer common pairs and triplets compared to \emph{gain} (e.g., \emph{loss} and \emph{call to action} appeared in just 5\% of texts).}

\takeaway{\changed{Framing was a relatively rare occurrence in the dataset, though predominantly present in phishing emails, wherein \emph{gain} was invoked $1.5\times$ more often than \emph{loss}.}}

\subsubsection{Emotional Salience}
\changed{We used VADER’s compound sentiment score ($E$, wherein $E\geq0.05$, , $E\leq-0.05$, and $-0.05<  E < 0.05$ denote positive, negative, and neutral sentiment, respectively) and LIWC’s positive and negative emotion word count metrics to measure sentiment. Overall, our dataset was slightly positive in terms of average compound sentiment ($\mu=0.23$) and with an average of 4.0 positive emotion words and 1.7 negative emotion words per text.}

\changed{In terms of specific text types, fake news contained the only negative average compound sentiment ($-0.163$), and right-leaning hyperpartisan news had the only neutral average compound sentiment ($0.015$); all other text types had, on average, positive sentiment, with phishing emails as the most positive text type ($0.635$). Left-leaning hyperpartisan news had the highest average positive emotion word count ($5.649$) followed by phishing emails ($4.732$), whereas fake news had the highest average negative word count ($2.892$) followed by left hyperpartisan news ($2.796$).}

We also analyzed whether emotional salience has indicative powers to predict the \changed{influence cues. Most influence cues and LIWC categories had an average positive sentiment, with \emph{liking} and \emph{gain} framing having the highest levels of positive emotion. \emph{Anxiety} and \emph{anger} (both LIWC categories) showed the only neutral sentiment, whereas \emph{admonition, blame/guilt}, and \emph{indignation} as the only negative sentiment (with the latter being the most negative out of all categories). Interestingly, items such as \emph{loss} framing and LIWC’s \emph{risk} both had positive sentiment.}


\takeaway{\changed{The dataset invoked an overall positive sentiment, with phishing emails containing the most positive average sentiment and fake news with the most negative average sentiment.}}

\subsubsection{\changed{Slant}}
\changed{The \emph{objective} and \emph{subjective} labels were present in 52\% and 64\% of all texts in the dataset, respectively. This $>50$\% frequency for both categories was present in all text types except phishing emails and IRA ads, where \emph{subjectivity} was approximately $2.5\times$ more common than \emph{objectivity}. The most subjective text type were IRA ads (77\%) and the most objective texts were fake news (72\%); inversely, the least objective texts were phishing emails (27\%) and least subjective were mainstream center news (58\%). 
}

\changed{More notably, there was an overlap between the slants, wherein 29\% of all texts contained both \emph{subjective} and \emph{objective} labels. This could reflect mixing factual (objective) statements with subjective interpretations of them. Nonetheless, \emph{objectivity} and \emph{subjectivity} were independent variables, $\chi^2 (4, N=2,998) = 72.0, p \approx 0$. The parings \emph{(objectivity, authority)} and \emph{(subjectivity, authority)} were the the top two most common pairs considering PoP and slant; these pairs occurred at nearly the same frequency within the dataset (48\% and 45\%, respectively). This pattern repeats itself for other (PoP, slant) pairings and triplets, insofar as \emph{(objectivity, subjectivity, authority)} is the third most commonly occurring triplet. When comparing just (PoP, slant) triplets, slant is present in $9/10$ top triplets, with \emph{(subjectivity, authority, commitment)} and \emph{(objectivity, authority, commitment)} as the two most common triplets (30\% and 27\%, respectively). 
}

\takeaway{\changed{\emph{Objectivity} and \emph{subjectivity} occurred over half of the dataset, with the latter was much more common in phishing emails and IRA ads, while the former was most common in fake news articles.}}

\subsubsection{\changed{Attribution of Blame \& Guilt}}
\changed{Twenty-nine percent of all texts contained the \emph{blame/guilt} label. Interestingly, nearly the same proportions of fake news (45.4\%) and right-leaning hyperpartisan news (45.0\%) were labeled with \emph{blame/guilt}, followed by left-leaning hyperpartisan news (38\%). Phishing emails, IRA ads, and mainstream center media used \emph{blame/guilt} at the lowest frequencies (ranging from 15\% to 25\%). }

\changed{\emph{Blame/guilt} was somewhat seen in the top 10 pairs with PoP, only pairing with \emph{authority} (4th most common pairing with 26\% frequency) and \emph{commitment} (6th most common, 18\%). However, \emph{blame/guilt} appeared more frequently amongst the top 10 triplets with PoP, co-occurring with \emph{authority, commitment, call to action,} and \emph{social admonition}.}

\takeaway{\changed{\emph{Blame/guilt} was disproportionately frequent for fake and hyperpartisan news, commonly co-occurring with \emph{authority} or \emph{commitment}.}}

\subsubsection{\changed{Emphasis}}
\changed{\emph{Emphasis} was used in nearly 35\% of all texts in the dataset. Among them, all news sources (fake, hyperpartisan, and mainstream) appeared with the smallest use of \emph{emphasis} (range: 17\% to 26\%). This follows as news (regardless of veracity) likely is attempting to purport itself as legitimate. On the other hand, phishing emails and IRA ads were both shared on arguably more informal environments of communication (email and social media), and were thus often found to use emphasis (over 54\% for both categories). Additionally, similar to previous analyses for other influence cues, \emph{emphasis} largely co-occurred with \emph{authority, commitment,} and \emph{call to action}.}

\takeaway{\changed{The use of emphasis was much more common in informal text typed (phishing emails and IRA social media ads), and less common in news-like sources (fake, hyperpartisan, or mainstream).}}

\subsubsection{LIWC Features of Influence}
We also explored whether LIWC features have indicative powers to predict the \changed{influence cues. Table~1 in Appendix~B shows that \emph{indignation} and \emph{admonition} had the highest average \emph{anxiety} feature, while \emph{liking} and \emph{gain} framing had the lowest. \emph{Indignation} also scored three times above the overall average for the \emph{anger} feature, as well as for \emph{sadness} (alongside \emph{blame/guilt}), whereas \emph{gain} had the lowest average for both \emph{anger} and \emph{sadness}. The \emph{reward} feature was seen most in \emph{liking} and in \emph{gain}, while \emph{risk} was slightly more common in \emph{loss} framing. The \emph{time} category had the highest overall average and was most common in \emph{blame/guilt}, while \emph{money} had the second largest overall average and was most common in \emph{loss}.}

\changed{We also saw that that left-leaning hyperpartisan news had the highest average \emph{anxiety}, \emph{sadness}, \emph{reward}, and \emph{time} counts compared to all text types, whereas right-leaning hyperpartisan news averaged slightly higher than left-leaning media only in the \emph{risk} feature. Note, however, that LIWC is calculated based on word counts and is therefore possibly biased towards longer length texts; it should thus be noted that while hyperpartisan left media had the highest averages for four of the seven LIWC features, hyperpartisan media also had the second largest average text length compared to other text types.}

\changed{For the Deceptive Texts Group, phishing emails had the largest \emph{risk} and \emph{money} averages over all text types, while averaging lowest in \emph{anxiety}, \emph{anger}, and \emph{sadness}. Fake news was highest overall in \emph{anger}, though it was slightly higher in \emph{anxiety}, \emph{sadness}, and \emph{time} compared to phishing emails and IRA ads. On the other hand, the IRA ads were lowest in \emph{reward}, \emph{risk}, \emph{time}, and \emph{money} compared to the its group.}

\changed{Lastly, mainstream center media had no LIWC categories in either high or low extremities—most of its average LIWC values were close to the overall averages for the entire dataset.}

\takeaway{\changed{LIWC influence features varied depending on the type of text. Left hyperpartisan news had the highest averages for four features (\emph{anxiety}, \emph{sadness}, \emph{reward}, and \emph{time}). Phishing evoked \textit{risk} and \textit{money}, while fake news evoked \textit{anger}.}}

\subsection{Lumen's Multi-Label Prediction}
\begin{table}[]
\centering
\caption{\changed{Evaluation metric results for different learning algorithms in detecting influence cues.}}
\label{tab:learning_results}
\begin{tabular}{@{}cccc@{}}
\toprule
\textbf{Algorithm} & \textbf{$F1$-macro} & \textbf{$F1$-micro} & \textbf{Overall Accuracy} \\ \midrule
{\bf Lumen}        & {\bf 58.30\%}       & {\bf 69.23\%}       & {\bf 72.43\%}      \\
Labeled-LDA        & 52.35\%             & 60.55\%             & 64.22\%            \\
LSTM               & 64.20\%             & 69.48\%             & 72.34\%             \\
Na\"{\i}ve         & 43.55\%             & 46.80\%             & 49.58\%             \\ \bottomrule
\end{tabular}
\end{table}

This section describes our results in evaluating Lumen's multi-label prediction using the dataset.
\changed{We compared Lumen’s performance against three other ML algorithms: Labeled-LDA, LSTM, and a naïve algorithm.} The former two learning algorithms and Lumen performed much better than the naïve algorithm, which shows that ML is promising for retrieval of influence cues in texts. \changed{From Table~\ref{tab:learning_results} we can see that Lumen's performance is as good as the state-of-the-art prediction algorithm LSTM in terms of $F1$-micro score and overall-accuracy (with $<$ 0.25\% difference between each metric).}
On the other hand, LSTM outperformed Lumen in terms of $F1$-macro, \changed{which is an unweighted mean of the metric for each labels, thus potentially indicating that Lumen underperforms LSTM in some labels although both algorithms share similar overall prediction result (accuracy). Nonetheless,} Lumen presented better interpretability than LSTM (discussed below).
Finally, both Lumen and LSTM presented better performance than Labeled-LDA in both $F1$-scores \changed{and accuracy, further emphasizing that additional features besides topic structures can help improve the performance of the prediction algorithm.}

To show Lumen's ability to provide better understanding to practitioners (i.e., interpretability), we trained it with our dataset and the optimal hyper-parameter values from grid search. After  training, Lumen provided both the relative importance of each input feature and the topic structure of the dataset without additional computational costs, which LSTM cannot provide because it operates as a black-box.


\begin{table}[h]
\centering
\caption{\changed{The top-five most important features for Lumen's prediction. \texttt{Topic} features are related to LDA topic modeling results.}}
\label{tab:important_features}
\begin{tabular}{@{}ccc@{}}
\toprule
\textbf{Input Feature} & \textbf{Importance} & \textbf{Keywords}                                                 \\ \midrule
Topic-1                & 0.073                       & \textit{account, bank, security, time}                    \\
Positive sentiment     & 0.071                       & N/A                                                               \\
Negative sentiment     & 0.070                       & N/A                                                               \\
Topic-8                & 0.065                       & \textit{report, share, billion, source,  profit} \\
Topic-2                & 0.062                       & \textit{black, people, trump, police, twitter}               \\ \bottomrule
\end{tabular}
\end{table}

Table~\ref{tab:important_features} shows the top-five important features in Lumen's prediction decision-making process. Among these features, \changed{two are related to sentiment, and the remaining three are topic features (related to bank account security, company profit report, and current events tweets)}, which shows the validity for the choice of these types of input features. 
\changed{Positive and negative sentiment had comparable levels of importance to Lumen, alongside the bank account security topic.}

\section{Discussion}
\label{sec:discussion}

In this paper, we posit that interventions to aid human-based detection of \changed{deceptive texts should leverage} a key invariant of these attacks: the application of influence cues in the text to increase its appeal to users. The exposure of these influence cues to users can potentially improve their decision-making \changed{by triggering their analytical thinking} when confronted with suspicious texts, which were not flagged as malicious via automatic detection methods. Stepping towards this goal, we introduced Lumen, a learning framework that combines topic modeling, LIWC, sentiment analysis, and ML to expose the following influence cues in deceptive texts: persuasion,  gain or loss framing, emotional salience, \changed{subjectivity or objectivity, and use of emphasis or attribution of guilt. 
Lumen was trained and tested on a newly developed dataset of 2,771 texts, comprised of purposefully deceptive texts, and hyperpartisan and mainstream news, all} labeled according to influence cues.

\subsection{Key Findings}
Most texts in the dataset applied \changed{between three and six influence cues}; we hypothesize that these findings may reflect the potential appeal or popularity of texts of moderate complexity. Deceptive or misleading texts constructed without any influence cues are too simple to convince the reader, while texts with too many influence cues might be far too long or complex, which are in turn more time-consuming to write (for attackers) and to read (for receivers). 

\changed{Most texts also applied \emph{authority}, which is concerning as it has been shown to be one of the most} impactful in user susceptibility to phishing studies~\cite{Oliveira2017dissect}. 
\changed{Meanwhile, \emph{reciprocation} was the least used principle at only 5\%; this may be an indication that \emph{reciprocation} does not lend itself well to be applied in text, as it requires giving something to the recipient first and expecting an action in return later. Nonetheless, \emph{reciprocation} was most common in IRA ads (10\%); these ads were posted on Facebook, and social media might be a more natural and intuitive location to give gifts or compliments.} 
We also found that the application of the PoP was highly imbalanced with \changed{\emph{reciprocation, indignation, social proof,} and \emph{admonition} each being applied less than 15\% the texts during the coding process.}

\changed{The least used influence cue were \emph{gain} and \emph{loss} framing, appearing in only 13\% and 7\% of all texts. Though Kahneman and Tversky~\cite{Kahneman1979-jz} posited that \emph{loss} is more impactful than the possibility of a gain, our dataset indicates that \emph{gain} was more prevalent than \emph{loss}. This is especially the case in phishing emails, wherein the framing frequencies increase to 41\% and 29\%;} 
this difference suggests that in phishing emails, attackers might be attempting to lure users to potential financial gain. \changed{We further hypothesize that phishing emails exhibited these high rates of framing because successful phishing survives only via a direct action from the user (e.g., clicking a link), which may therefore motivate attackers to implement framing as a key influence method. Phishing emails also exhibited the most positive average sentiment (0.635)  compared to other text types, possibly related to its large volume of \emph{gain} labels, which were also strongly positive in sentiment (0.568).}

\changed{Interestingly, texts varied among themselves in terms of influence cues even within their own groups. For example, within the Deceptive Texts Group, fake news used notably more \emph{authority}, \emph{objectivity}, and \emph{blame/guilt} compared to phishing emails and IRA ads, and was much lower in sentiment compared to the latter two. Though phishing emails and IRA ads were more similar, phishing was nonetheless different in its use of higher positive sentiment, \emph{gain} framing, \emph{scarcity}, and lower \emph{blame/guilt}. This was also evident within the Hyperpartisan News Group---while right-learning news had a higher frequency of \emph{commitment}, \emph{call to action}, and \emph{admonition} than left-learning news, the opposite was also true for \emph{liking}, \emph{reciprocation}, and \emph{scarcity}. Even comparing among all news types (fake, hyperpartisan, and mainstream), this diversity of influence cues still prevailed, with the only resounding agreement in a relative lack of use of \emph{emphasis}. This diversity across text types gives evidence of the highly imbalanced application of influence cues in  real deceptive or misleading campaigns.}

We envision the use of Lumen (and ML methods in general) to expose influence cues as a promising direction for \changed{application tools to aid human detection of cyber-social engineering and disinformation. Lumen presented a comparable performance compared to LSTM} in terms of the $F1$-micro. Lumen's interpretability can allow a better understanding of both the dataset and the decision-making process Lumen undergoes, consequently providing invaluable insights for feature selection.

\subsection{Limitations \& Future Work}
\label{sec:limitations}

\textbf{\changed{Dataset.}} One of the limitations of our work is that the dataset is unbalanced. For example, our coding process revealed that some influence (e.g., \emph{authority}) were disproportionately more prevalent than others (e.g., \emph{reciprocation}, framing). Even though an unbalanced dataset is not ideal for ML analyses, we see this as part of the phenomenon. Attackers  and writers might find it more difficult to construct certain concept via text, thus favoring other more effective and direct influence cues such as \emph{authority}. 
\changed{Ultimately, our dataset is novel in that each of the nearly 3K items were coded according to 12 different variables; this was a time-expensive process and we shall test the scalability of Lumen in future work. Nevertheless, we plan to alleviate this dataset imbalance in our future work by curating a larger, high-quality labeled dataset by reproducing our coding methodology,} and/or with the generation of synthetic, balanced datasets. 
\changed{Though we predict that a larger dataset will still have varying proportions of certain influence cues, it will facilitate machine learning with a larger volume of data points.}

\changed{Additionally, our dataset is U.S.-centric, identified as a limitation in some prior work (e.g., \cite{Kalsnes2021-pg, Fletcher2018-jo, Newman2019-zh}). All texts were ensured to be in the English language and all three groups of data were presumably aimed at an American audience. Therefore, we plan future work to test Lumen in different cultural contexts.}


\changed{\textbf{ML Framework.}} Lumen, as a learning framework, has three main limitations. First, although the two-level architecture provides high degree of flexibility and is general enough to include other predictive features in the future, it also introduces complexity and overhead because tuning the hyper-parameters and training the model will be more computationally expensive. 

Second, topic modeling, a key component of Lumen, generally requires a large number of documents of a certain length (usually thousands of documents and hundreds of words in each document, such as a collection of scientific paper abstracts) for topic inference. This will limit Lumen's effectiveness on short texts or when the training data is limited. 

Third, some overlap between the LIWC influence features and emotional salience might exist (e.g., the \emph{sad} LIWC category may correlate with the negative emotional salience), which may negatively impact the prediction performance of the machine learning algorithm used in Lumen. In other words, correlation of input features makes machine learning algorithms hard to converge in general.



\section{Conclusion}
\label{sec:conclusion}
In this paper, we introduced Lumen, a learning-based framework to expose \changed{influence cues in text by combining} topic modeling, LIWC, sentiment analysis, and machine learning in a two-layer hierarchical architecture. Lumen was \changed{trained and tested with a newly developed dataset of 2,771 total texts manually} labeled according to the influence cues applied to the text. Quantitative analysis of the dataset showed that \changed{\textit{authority} was the most prevalent influence cue, followed by \textit{subjectivity} and \textit{commitment}; \textit{gain} framing was most prevalent in phishing emails, and use of emphasis commonly occurred in fake, partisan, and mainstream news articles.}
Lumen presented \changed{comparable performance with LSTM} in terms of $F1$-micro score, but better interpretability, providing insights of feature importance. Our results highlight the promise of ML to expose influence cues in text with the goal of \changed{application in tools to improve the accuracy of human detection of cyber-social engineering threats, potentially triggering users to think analytically.}
We advocate that the next generation of interventions to mitigate deception expose influence to users, complementing automatic detection to address \changed{new deceptive campaigns} and improve user decision-making when confronted with potentially suspicious text.


\ifCLASSOPTIONcompsoc
  \section*{Acknowledgments}
\else
  \section*{Acknowledgment}
\fi
The authors would like to thank the coders for having helped with the labeling of the influences cues in our dataset. This work was support by the University of Florida Seed Fund award P0175721 and by the National Science Foundation under Grant No 2028734. This material is based upon work supported by (while serving at) the National Science Foundation.

\ifCLASSOPTIONcaptionsoff
  \newpage
\fi



%
%
%

\bibliographystyle{IEEEtran}
\bibliography{references}

%

\vskip -2\baselineskip plus -1fil

\begin{IEEEbiographynophoto}{Hanyu~Shi}
	was a Postdoc Researcher under Dr. Daniela Oliveira at the University of Florida during this work. He received his PhD degree with a special focus on topic modeling algorithms in the department of Chemical and Biological Engineering at Northwestern University, US. His current research interests lie in the area of NLP, statistic inference, information retrieval, etc.
\end{IEEEbiographynophoto}

\vskip -2\baselineskip plus -1fil

\begin{IEEEbiographynophoto}{Mirela~Silva}
	is currently pursuing her PhD at UF under the Dr. Oliveira. Her research interests include interdisciplinary computer privacy, focused heavily on the intersection of cyber deception and abuse through the lens of interindividual differences.
\end{IEEEbiographynophoto}

\vskip -2\baselineskip plus -1fil

\begin{IEEEbiographynophoto}{Daniel~Capecci}
	is currently pursuing his Ph.D. under Dr. Damon Woodard at the UF. His current research interests include computer vision and machine learning for hardware assurance, large network analysis and deepfake detection.
\end{IEEEbiographynophoto}

\vskip -2\baselineskip plus -1fil

\begin{IEEEbiographynophoto}{\changed{Luiz~Giovanini}}
	\changed{is a Postdoc Researcher under Dr. Olivera at UF. He is also an assistant professor at the Pontifical Catholic University of Parana (PUCPR), Brazil. He received his PhD degree in Computer Science with focus on applied machine learning from PUCPR in Brazil. His current research interests include machine learning for applications in disinformation and cyber security.}
\end{IEEEbiographynophoto}

\vskip -2\baselineskip plus -1fil

\begin{IEEEbiographynophoto}{\changed{Lauren~Czech}}
	\changed{is pursuing a Ph.D. under Dr. Oliveira. Her current research interests include leveraging ML to increase animal welfare.}
\end{IEEEbiographynophoto}

\vskip -2\baselineskip plus -1fil

\begin{IEEEbiographynophoto}{\changed{Juliana~Fernandes}}
	\changed{is an Assistant Professor of Advertising at UF. She received her BA in Journalism from Universidade do Vale do Rio dos Sinos, Brazil, MA and PhD degrees in Mass Communication from UF. Her current research interests include uses and effects of negative information in persuasive communication.}
\end{IEEEbiographynophoto}

\vskip -2\baselineskip plus -1fil

\begin{IEEEbiographynophoto}{Daniela~Seabra~Oliveira}
	is an Associate Professor in the Department of Electrical and Computer Engineering at UF. She received her BS and MS degree in Computer Science from the Federal University of Minas Gerais in Brazil. She received her PhD in Computer Science from UC-Davis, where she specialized in security and operating systems. Her current research interests are human factors in cyber security, phishing, and disinformation.
\end{IEEEbiographynophoto}


%

%

\newpage

\appendices

\section{Codebook}
\label{appendix:codebook}

\subsection*{Persuasion}
\textit{Persuasion constitutes a series of influence principles based on Robert Cialdini's work, split into the following categories:}
\begin{enumerate}
    \item Authority or Expertise/Source Credibility
    \item Reciprocation
    \item Commitment (sub-categories: Indignation, Call to Action)
    \item Liking
    \item Scarcity/Urgency/Opportunity
    \item Social Proof (sub-category: Admonition)
\end{enumerate}

\textbf{Authority or Expertise/Source Credibility.} 
Humans tend to comply with requests made by figures of authority and/or with expertise/credibility. The text can include: 
\begin{itemize}
    \item Literal authority (e.g., law enforcement personnel, lawyers, judges, politicians)
    \item Reputable/credible entity that could exert some power over people  (e.g., a bank)
    \item Indirect authority (especially a fictitious company/person) that builds a setting of authority
\end{itemize}

Examples:
\begin{itemize}
    \item ``\textit{Tupac Shakur was indeed not just one of the greatest rappers of all time but a worldly icon} whose status in hip-hop culture can never be replaced. His revolutionary knowledge mixed with street experience made him powerful unstoppable force that spoke to the hearts of millions of people.''
    \item ``According to \textit{data from Mapping Police Violence}''
    \item ``\textit{Autopsy} says''
    \item ``\textit{Fox \& Friends hosts} declare''

\end{itemize}


\textbf{Reciprocation.}
Humans tend to repay, in kind, what another person has provided them.
Text might first give/offer something, expecting that the person/user will reciprocate. Even if the person does not reciprocate, s/he will still keep the "gift." Therefore, if the user thinks they received a gift, they may reciprocate the kindness (and may only find out later that the "gift" was fake).

Example:
\begin{itemize}
    \item ``Aww! Because you need such a cutie on your timeline!''
\end{itemize}


\textbf{Commitment.}
Once humans have taken a stand, they will feel pressured to behave in line with their commitment.
Text leverages a role assumed by the target and their commitment to that role. Petitions and donations/charity (gun control, animal abuse, children's issue, political issues), or political affiliations engagements.

Examples:
\begin{itemize}
    \item ``But let's remember Tupac and his ability to question the social order. Changes, one of his popular songs, \textit{asks everyone to change their lifestyles for better society. He always asked people to share with each other and to learn to love each other}.''
    \item ``Patriotism comes from your heart... follow its dictates and don't live a false life. Join!''
    \item ``We will stand for our right to keep and bear arms!''
    \item ``Black Matters''

\end{itemize}


\textbf{Indignation.}
Still within the definition of \textit{commitment}, text employing \textit{indignation} will also focus on anger or annoyance provoked by what is perceived as unfair treatment, unjust, unworthy, mean. 

Examples:
\begin{itemize}
    \item ``Why should we be a target for police violence and harassment?'' 
    \item ``Why the pool party in Georgia is a silent story? Why the police was not aware of a large party? Why this story has no national outrage? Is it ok when a black teenager dies?''
    \item ``Obama never tried to protect blacks from police pressure''
\end{itemize}


\textbf{Call to Action.}
Still within the definition of \textit{commitment}, ads/text employing a \textit{call to action} will represent an exhortation or stimulus to do something in order to achieve an aim or deal with a problem.
A piece of content intended to induce a viewer, reader, or listener to perform a specific act, typically taking the form of an instruction or directive (e.g., \textit{buy now} or \textit{click here}).

Examples:
\begin{itemize}
    \item ``Stop racism! We all belong to ONE HUMAN RACE.'' 
    \item ``We really can change the world if we stay united''
    \item ``We can be heard only when we stand together''
    \item ``White House must reduce the unemployment rates of black population''
    \item ``If this is a war against police - we're joining this war on the cop's side!''
    \item ``If we want to stop it, we should fight as our ancestors did it for centuries.''
\end{itemize}


\textbf{Liking.}
Humans tend to comply with requests from people they like or with whom they share similarities. Forms of \textit{liking}:
\begin{itemize}
    \item Physical attractiveness: Good looks suggest other favorable traits, i.e. honesty, humor, trustworthiness
    \item Similarity: We like people similar to us in terms of interests, opinions, personality, background, etc.
    \item Compliments: We love to receive praises, and tend to like those who give it
    \item Contact and Cooperation: We feel a sense of commonality when working with others to fulfill a common goal
    \item Conditioning and Association: We like looking at models, and thus become more favorable towards the cars behind them
\end{itemize}
May also come in the form of establishing a familiarity or rapport with the object of liking

Example:
\begin{itemize}
    \item ``What a beautiful and intelligent child she is. How magnificent is her mind…''
\end{itemize}


\textbf{Scarcity/Urgency/Opportunity.}
Opportunities seem more valuable when their availability is limited
Text can leverage this principle by tricking/asking an Internet user into clicking on a link to avoid missing out on a ``once-in-a-lifetime'' opportunity; creates a sense of urgency.

Examples:
\begin{itemize}
    \item ``Is it time to call out the national guard?'' (Urgency)
    \item ``Free Figure's Black Power Rally at VCU'' (Opportunity)
    \item ``CLICK TO GET LIVE UPDATES ON OUR PAGE'' (Opportunity)
\end{itemize}


\textbf{Social Proof.}
People tend to mimic what the majority of people do or seem to be doing.
People let their guard and suspicion down when everyone else appears to share the same behaviors and risks. In this way, they will not be held solely responsible for their actions (i.e., herd mentality). The actions of the group drive the decision making process.

Examples:
\begin{itemize}
    \item ``More riots are coming this summer''
    \item ``America is deceased. Islamic terror has penetrated our homeland and now spreads at a threw. Remember Victims Of Islamic Terror''
\end{itemize}


\textbf{Admonition.}
Within the definition of \textit{social proof}, \textit{admonition} pertains to texts that may include the following:
\begin{itemize}
    \item Caution, advise, or counsel against something. 
    \item Reprove or scold, especially in a mild and good-willed manner: \textit{The teacher admonished him about excessive noise.}
    \item Urge to a duty or admonish them about their obligations.
\end{itemize}

Examples:
\begin{itemize}
    \item ``More riots are coming this summer''
    \item ``America is deceased. Islamic terror has penetrated our homeland and now spreads at a threw Remember Victims Of Islamic Terror''
\end{itemize}

\subsection*{Slant}
\textit{Slant encompasses subjectivity and objectivity.}

\textbf{Subjectivity.}
Subjective sentences generally refer to personal opinion, emotion or judgment. The use of popular adverbs (e.g, very, actually), upper case, exclamation and interrogation marks, hash tags, indicates subjectivity.

Examples:
\begin{itemize}
    \item ``I doubt that it's true''
    \item ``A beautiful message was seen on the streets of the capitol,'' 
    \item ``A timely message for today.'' 
    \item ``No matter what Defense Secretary or POTUS are saying they don't fool me with promises of gay military equality as key to the nation's agenda.''
    \item ``This is something that America has a serious issue with - RACISM!'' ``Is it time to call out the national guard?'' 
    \item "This makes me ANGRY!''
\end{itemize}

\textbf{Objectivity.}
Objective sentences refers to factual information, based on evidence, or when evidence is presented. May or may not include statistics. 

Examples:
\begin{itemize}
    \item ``It has been discovered that''
    \item ``According to data from Mapping Police Violence''
    \item ``The McKinney Police Department, Chief Of Police Greg Conley said''
\end{itemize}

\subsection*{Framing}
\textit{Gain/Loss Framing refers to the presentation of a message (e.g., health message, financial options, advertisement etc.) as implying a possible gain (e.g., refer to possible benefits of performing a behavior) vs. implying a possible loss (e.g., refer to the costs of not performing a behavior).}

\textbf{Gain.}
People are likely to act in ways that benefit them in some way. A reward will increase the probability of a behavior
A promise that a product (or something else) can provide some form of self-improvement or benefit to the user. This product can come in the form of an ad, job offer, joining a group, etc.

Examples:
\begin{itemize}
    \item ``Think about the benefits of recycling."
    \item ``Think about what you can gain if you join."
\end{itemize}

\textbf{Loss.}
People are likely to act in ways that reduce loss/harm to them. Avoiding loss will increase the probability of a behavior
A promise that a product (or something else) can help avoid some behavior/outcome. 

Examples:
\begin{itemize}
    \item ``Think about the costs of recycling."
    \item ``Think about what you can lose if you don't join."
\end{itemize}

\subsection*{Attribution of Blame/Guilt}
\textit{When the text references an ``another'' (who/what) for the wrong/bad things happening. ``Who'' can be a person, organization, etc., and ``what'' can be a cause, object, etc.}

Example:
\begin{itemize}
    \item ``...Hillary is a Satan, and her crimes and lies had proved just how evil she is.''
\end{itemize}

\subsection*{Emphasis}
\textit{Emphasis refers to the use of all caps text, several exclamation points, several question marks, or anything used to call attention.}

Example:
\begin{itemize}
    \item ``Our women are the most powerful!!''
    \item ``LATIN WOMEN CAN DO THINGS TO MEN WITH THERE EYES''
\end{itemize}



\section{LIWC Emotional Features Summary}
\label{appendix:liwc_features}

\vspace{-4em}
\begin{table*}[]
\centering
\caption{Breakdown of LIWC emotional features for each influence cue and text type; highest values for each LIWC category (columns) per influence cue and text type are highlighted bold text.}
\label{tab:LIWC_influence}
\begin{tabular}{@{}cccccccc@{}}
\toprule
\textbf{\begin{tabular}[c]{@{}c@{}}
Influence Cue or\\Text Type\end{tabular}} &
  \textbf{Anxiety} &
  \textbf{Anger} &
  \textbf{Sadness} &
  \textbf{Reward} &
  \textbf{Risk} &
  \textbf{Time} &
  \textbf{Money} \\ \midrule
\textit{Authority}                & 0.23          & 0.69           & 0.36          & 1.69          & 0.78           & 6.68          & 2.21           \\
\textit{Commitment}               & 0.20          & 0.67           & 0.28          & 1.61          & 0.78           & 5.97          & 1.87           \\
\textit{Call to Action}           & 0.18          & 0.60           & 0.26          & 1.59          & 0.94           & 5.75          & 2.19           \\
\textit{Indignation}              & \textbf{0.34} & \textbf{1.62}  & \textbf{0.46} & 1.49          & 0.91           & 6.06          & 0.88           \\
\textit{Liking}                   & 0.13          & 0.34           & 0.24          & \textbf{1.84} & 0.57           & 5.87          & 1.67           \\
\textit{Reciprocation}            & 0.19          & 0.32           & 0.36          & 1.53          & 0.59           & 5.00          & 1.14           \\
\textit{Scarcity}                 & 0.18          & 0.51           & 0.30          & 1.67          & 1.01           & 5.84          & 2.73           \\
\textit{Social Proof}             & 0.26          & 0.69           & 0.37          & 1.61          & 0.79           & 5.81          & 1.64           \\
\textit{Admonition}               & \textbf{0.34} & 1.14           & 0.39          & 1.52          & 1.09           & 6.43          & 1.52           \\
\textit{Emphasis}                 & 0.20          & 0.51           & 0.26          & 1.43          & 0.87           & 5.51          & 2.27           \\
\textit{Blame/guilt}              & 0.30          & 1.22           & \textbf{0.47} & 1.47          & 1.00           & \textbf{6.85} & 1.60           \\
\textit{Gain framing}             & 0.12          & 0.23           & 0.24          & \textbf{1.89} & 1.09           & 5.50          & 3.42           \\
\textit{Loss framing}             & 0.20          & 0.41           & 0.37          & 1.70          & \textbf{1.16}  & 5.78          & \textbf{3.73}  \\
\textit{Objectivity}              & 0.23          & 0.70           & 0.36          & 1.60          & 0.74           & 6.71          & 2.31           \\
\textit{Subjectivity}             & 0.22          & 0.61           & 0.32          & 1.60          & 0.78           & 6.05          & 1.96           \\ \midrule
\textit{Fake News}                & 0.25         & \textbf{1.08} & 0.35         & 1.35         & 0.80          & 7.51         & 1.56          \\
\textit{(News) Center}            & 0.19         & 0.34          & 0.39         & 1.54         & 0.60          & 6.20         & 3.01          \\
\textit{(News) Left} &
  \textbf{0.33} &
  1.03 &
  \textbf{0.46} &
  \textbf{2.26} &
  0.74 &
  \textbf{8.85} &
  1.14 \\
\textit{(News) Right}             & 0.25         & 1.01          & 0.30         & 1.69         & 0.79          & 6.66         & 0.88          \\
\textit{Phishing Email}           & 0.06         & 0.05          & 0.15         & 1.63         & \textbf{1.21} & 4.96         & \textbf{4.27} \\
\textit{IRA Ads}                  & 0.12         & 0.55          & 0.18         & 0.65         & 0.54          & 2.34         & 0.56          \\ \midrule
\textit{\textbf{Average}} & 0.20         & 0.59          & 0.32         & 1.56         & 0.75          & 6.02         & 2.11          \\ \bottomrule
\end{tabular}
\end{table*}






\end{document}